\documentclass{article}

     \PassOptionsToPackage{numbers, compress}{natbib}

     \usepackage[final]{neurips_2019}

\usepackage[utf8]{inputenc} 
\usepackage[T1]{fontenc}    
\usepackage{hyperref}       
\usepackage{url}            
\usepackage{booktabs}       
\usepackage{amsfonts}       
\usepackage{nicefrac}       
\usepackage{microtype}      
\usepackage{graphicx}
\usepackage{subcaption}
\usepackage{pgfplots}
\usepackage{lipsum}
\usepackage{placeins}
\usepackage{listings}
\usepackage{color}

\definecolor{mygreen}{rgb}{0,0.6,0}
\definecolor{mygray}{rgb}{0.5,0.5,0.5}
\definecolor{mymauve}{rgb}{0.58,0,0.82}

\usepackage[toc,page]{appendix}
\usepackage{courier}

\pgfplotsset{compat=1.14}

\title{The PlayStation Reinforcement Learning Environment (PSXLE)}

\author{%
  Carlos Purves, Cătălina Cangea, Petar Veličković \\
  Department of Computer Science and Technology\\
  University of Cambridge \\
  \texttt{\{cp614, catalina.cangea, petar.velickovic\}@cst.cam.ac.uk} \\
}

\begin{document}

\maketitle

\begin{abstract}
    We propose a new benchmark environment for evaluating Reinforcement Learning (RL) algorithms: the \emph{PlayStation Learning Environment} (PSXLE), a PlayStation emulator modified to expose a simple control API that enables rich game-state representations. We argue that the PlayStation serves as a suitable progression for agent evaluation and propose a framework for such an evaluation. We build an action-driven abstraction for a PlayStation game with support for the \emph{OpenAI Gym} interface and demonstrate its use by running \emph{OpenAI Baselines}.
\end{abstract}

\section{Introduction}

\emph{Reinforcement Learning} (RL) describes a form of machine learning in which an agent learns how to interact with an \emph{environment} through the acquisition of \emph{rewards} that are chosen to encourage good behaviours and penalise harmful ones. The environment is described to the agent at each point in time by a \emph{state encoding}. A well-trained agent should use this encoding to select its next \emph{action} as one that maximises its long-term cumulative reward. This model of learning has proved effective in many real-world environments, including in self-driving cars~\cite{DrivingStudy}, traffic control~\cite{TrafficRL}, advertising~\cite{BiddingStudy} and robotics~\cite{RoboticStudy}.

An important advance in RL research came with the development of \emph{Deep Q-Networks} (DQN), in which agents utilise deep neural networks to interpret complex state spaces. Increased attention towards RL in recent years has led to further advances such as Double DQN~\cite{DoubleQ}, Prioritised Experience Replay~\cite{PExperienceReplay} and Duelling Architectures~\cite{DuellingArch} bringing improvements over DQN. Policy-based methods such as A3C have brought further improvements~\cite{AsyncMethods} and asynchronous methods to RL. 

In order to quantify the success of these learning algorithms and to demonstrate improvements in new approaches, a common evaluation methodology is needed. Computer games are typically used to fulfil this role, providing practical advantages over other types of environments: episodes are \emph{reproducible}, due to the lack of uncontrollable stochasticity; they offer comparatively low-dimensional state encodings; and the notion of a `score' translates naturally into one of a `reward'. The use of computer games also serves a more abstract purpose: to court public interest in RL research. Describing the conclusions of research in terms of a player's achievement in a computer game makes the work more approachable and improves its comprehensibility, as people can utilise their own experience of playing games as a baseline for comparison.

In 2015, Mnih \emph{et al.}~\cite{NatureDeepMind} used the Atari-2600 console as an environment in which to evaluate DQN. Their agent \emph{outperformed} human expert players in 22 out of the 49 games that were used in training. In 2016, OpenAI announced OpenAI Gym~\cite{OAIRelease}, which allows researchers and developers to interface with games and physical simulations in a standardised way through a Python library. Gym now represents the de-facto standard evaluation method for RL algorithms~\cite{GymInfo}. It includes, amongst others~\cite{GymENVS}, several Atari-2600 games which utilise Arcade Learning Environment (ALE)~\cite{ALE} to interface with a console emulation.

One of the most important considerations in developing successful RL methods in complex environments is the choice of \emph{state encoding}. This describes the relationship between the state of an environment and the format of the data available to the agent. For a human playing a game, `state' can mean many things, including: the position of a character in a world, how close enemies are, the remaining time on a clock or the set of previous actions. While these properties are easy for humans to quantify, RL environments usually do not encode them explicitly for two reasons. Firstly, doing so would simplify the problem too much, permitting reliance on a human understanding of the environment---something which should ideally be approximated through learning. Secondly, it does not allow agents to generalise, since each game's state will be described by different properties. Rather, game-based RL environments typically consider the `state' to be an element of some common \emph{state space}. Common examples of such spaces are the set of a console's possible display outputs or its RAM contents.

Until now, RL research has seen little exploration of the use of sound effects in state encodings. This is clearly not due to a lack of methods for processing audio data; there is substantial research precedent in the areas of speech recognition~\cite{SPEECH}, audio in affective computing~\cite{AFFECT} and unsupervised approaches to music genre detection~\cite{GENREDETECT}. A discussion about richer state encodings is particularly pertinent given the success of existing RL approaches within conventional environments. A significant gap exists between the richness and complexity of such environments and those representing the eventual goal of RL: real-world situations with naturally many-dimensional state spaces.

To help narrow this gap, this paper introduces the \emph{PlayStation Reinforcement Learning Environment} (PSXLE): a toolkit for training agents to play Sony PlayStation 1 games. PSXLE is designed to follow from the standard set by ALE and enable RL research using more complex environments and state encodings. It increases the complexity of the games that can be used within a framework such as OpenAI Gym, due to the significant hardware differences between the consoles. PSXLE utilises this additional complexity by exposing raw audio output from the console alongside RAM and display contents. We implement an \emph{OpenAI Gym} interface for the PlayStation game \emph{Kula World} and use \emph{OpenAI Baselines} to evaluate the performance of two popular RL algorithms with it.

\section{PlayStation}

\begin{figure}[h]
    \centering
    \includegraphics[width=8cm]{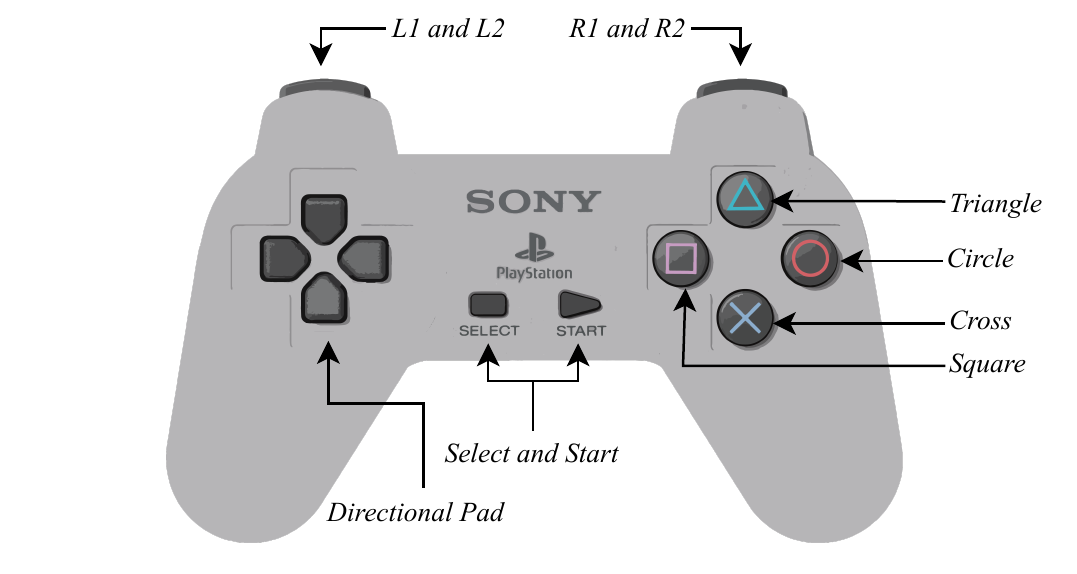}
    \caption{The PlayStation controller, highlighting its 14 buttons.}
    \label{fig:controller}
\end{figure}

The Sony PlayStation 1 (sometimes \emph{PSX} or \emph{PlayStation}) is a console first released by Sony Computer Entertainment in 1994. It has 2 megabytes of RAM, 16.7 million displayable colours and a 33.9 MHz CPU, which contrasts with the Atari-2600's 128 bytes of RAM, 128 displayable colours and a 1.19MHz CPU. Since its launch, the number of titles available for the PlayStation has grown to almost 8000 worldwide, more than the 500 that are available for the Atari-2600 console. PlayStation games are controlled using a handheld controller, shown in Figure \ref{fig:controller}. The controller has 14 buttons, with \texttt{Start} typically used to pause a game.

\section{Implementation}

\begin{figure}[t]
    \centering
    \includegraphics[width=.6\textwidth,keepaspectratio]{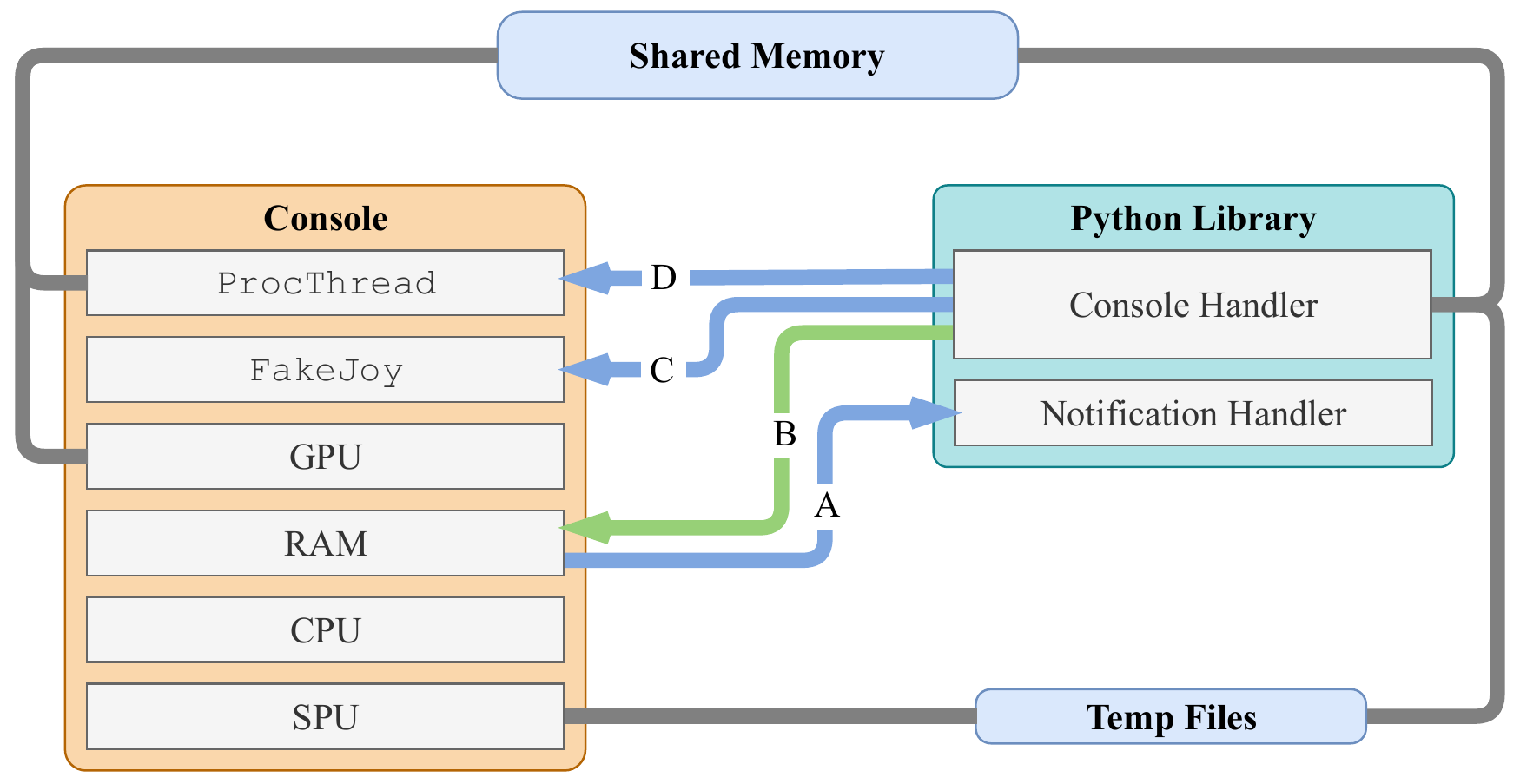}
    \caption{A visualisation of the Inter-Process Communication used in PSXLE. Pipes are coloured green, FIFO queues are coloured blue and Unix \texttt{fopen} calls are coloured grey.
    \textbf{A} is used to notify the PSXLE Python library that parts of memory have changed and that events like \texttt{load\_state} and \texttt{save\_state} have completed.
    \textbf{B}, representing standard input \texttt{stdin}, is used to communicate which regions of memory the console should watch.
    \textbf{C} is used to send simulated button presses to the console.
    \textbf{D} sends instructions to the console, such as loading and saving state or loading an ISO.}
    \label{fig:pipelayout}
\end{figure}

\begin{figure}[h]
    \centering
    \includegraphics[width=\textwidth]{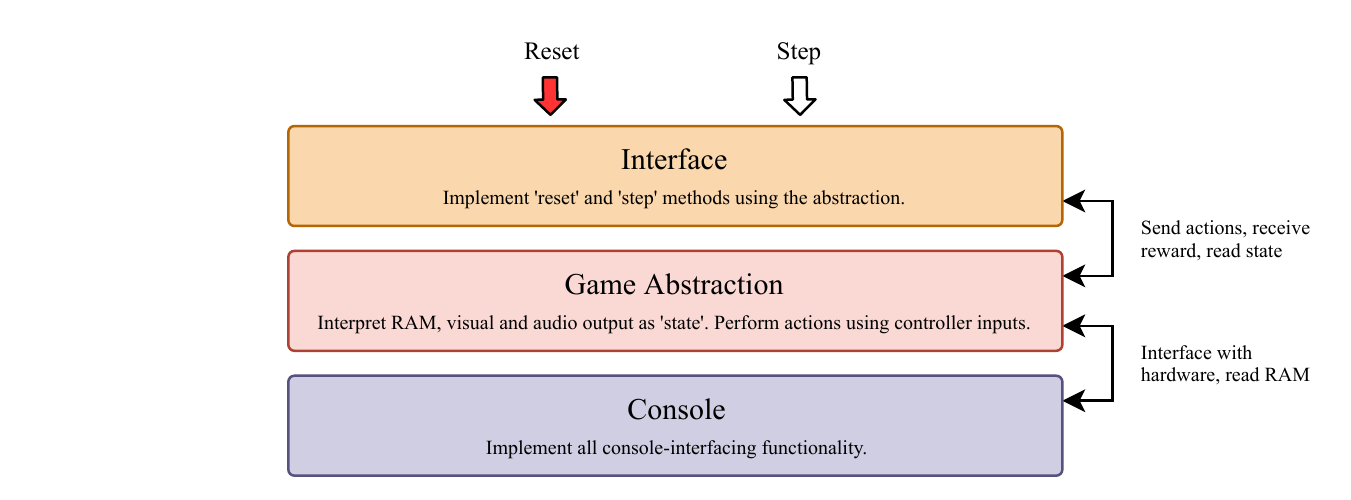}
    \caption{The proposed architecture for PlayStation RL environments includes three components: the \textbf{interface}, which allows agents to perform actions and receive feedback; the \textbf{game abstraction}, which translates game actions into controller inputs and console state (visual, audio and RAM) into state representations; and the \textbf{console}, which handles communication with the emulator.}
    \label{fig:envarch}
\end{figure}

PSXLE is built using a fork of \emph{PCSX-R}\footnote{Available at \url{https://github.com/pcsxr/PCSX-Reloaded/}}, an open-source emulator created in 2009. We made modifications to the source of PCSX-R by adding simple Inter-Process Communication (IPC) tools---the structure of which is shown in Figure \ref{fig:pipelayout}---to simulate controller inputs and read console output through an external interface. The PSXLE Python library uses these tools to provide a simple, game-agnostic PlayStation console interface. To specialise the environment to a certain game and implement an interface such as OpenAI Gym, a customised \emph{environment stack} can be created that abstracts console-level functions across two layers. Figure \ref{fig:envarch} visualises the structure of such a stack. Specialisation to individual games occurs within the `game abstraction' component, which uses the \emph{Console API} to translate game \emph{actions} (such as `move forwards') into console functions (such as `press up').

The Console API supports four primary forms of interaction:

\begin{itemize}
    \item General:
    \begin{itemize}
        \item \texttt{run} and \texttt{kill} control the executing process of the emulator;
        \item \texttt{freeze} and \texttt{unfreeze} will freeze and unfreeze the emulator's execution, respectively;
        \item \texttt{speed} is a property of \texttt{Console} which, when set, will synchronously set the speed of execution of the console, expressed as a percentage relative to default speed.
    \end{itemize}
    \item Controller:
    \begin{itemize}
        \item \texttt{hold\_button} and \texttt{release\_button} simulate a press down and release of a given controller button---referred to here as \emph{control events};
        \item \texttt{touch\_button} holds, pauses for a specified amount of time and then releases a button;
        \item \texttt{delay\_button} adds a (millisecond-order) delay between successive control events
    \end{itemize} 
    \item RAM:
    \begin{itemize}
        \item \texttt{read\_bytes} and \texttt{write\_byte} directly read from and write to console memory;
        \item \texttt{add\_memory\_listener} and \texttt{clear\_memory\_listeners} control which parts of the console's memory should have asynchronous listeners attached when the console runs;
        \item \texttt{sleep\_memory\_listener} and \texttt{wake\_memory\_listener} tell the console which listeners are active.
    \end{itemize}
    \item Audio/Visual:
    \begin{itemize}
        \item \texttt{start\_recording\_audio} and \texttt{stop\_recording\_audio} control when the console should record audio and when it should stop;
        \item \texttt{get\_screen} synchronously returns an \texttt{np.array} of the console's instantaneous visual output.
    \end{itemize}
\end{itemize}

Example usage of the PSXLE interface can be found in Appendix \ref{apdx:usage}.

\section{Game abstraction}

OpenAI Gym environments expose three methods: \texttt{reset}, which restarts an episode and returns the initial state; \texttt{step}, which takes an action as an argument and performs it within the environment; and \texttt{render}, which renders the current state to a window, or as text.

The \texttt{step} function takes an integer representing an action and returns a tuple containing: \texttt{state}, which is the value of the state of the system after an action has taken place; \texttt{reward}, which gives the reward gained by performing a certain action; \texttt{done}, which is a Boolean value indicating whether the episode has finished; and \texttt{info}, which gives extra information about the environment. Gym requires that these methods return synchronously. There are two possible approaches to deriving this synchrony with PlayStation games.

\begin{figure}
    \centering
    \begin{subfigure}[t]{\textwidth}
      \centering
      \includegraphics[width=\textwidth,keepaspectratio]{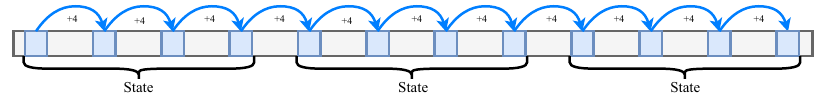}
      \caption{A visualisation of frame skip and frame stacking. In~\cite{NatureDeepMind}, only every fourth frame is considered and of those, every four frames are combined (stacked) into a single state representation. New frames are requested by the Python library upon each action, making this approach synchronous.}
    \end{subfigure} \\
    \begin{subfigure}[t]{\textwidth}
      \centering
      \includegraphics[width=\textwidth,keepaspectratio]{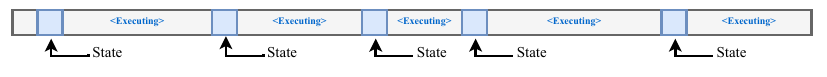}
      \caption{An asynchronous approach to frame skipping. In environments where actions are long or have variable length, the state transition occurs asynchronously. The transition ends once the immediate effects of the associated action have ceased.}
    \end{subfigure}
    \caption{A comparison of the state-delimiting techniques used in~\cite{NatureDeepMind} and those used in this paper.}
    \label{fig:enctechs}
\end{figure}

Firstly, the environment could exercise granular control over the execution of the console, choosing how many frames to \emph{skip} for each move. This approach is common and is used to implement the Atari-2600 environments in Gym. In cases where a simple snapshot of the environment would leave ambiguity (such as when the motion of an object could be in several directions), consecutive frames may be \emph{stacked} to produce a corresponding state encoding. Frame skipping works well for simple games, but is not always suitable if moves can take a variable amount of time to finish. If the skip is less than the number of frames a move takes to finish, the agent may choose its next action \emph{before} a previous move has finished. In many games, this would lead to the chosen action not being completed properly. If the environment was to skip significantly more frames than required, the agent would unnecessarily incur a delay in making moves.

A second approach would be to allow the console to run asynchronously for the duration of each move, with an associated \emph{condition} that signifies the move being over. For example, a move that involves collecting a valuable item might be `finished' once a player's score has changed. To the agent, the move would \emph{begin} when \texttt{step} was called and \emph{end} as soon as the score had increased. This approach is easy to implement with PSXLE, using memory listeners that respond to changes in RAM. These approaches are contrasted in Figure \ref{fig:enctechs}.

\section{Kula World}

\emph{Kula World} is a game developed by Game Design Sweden A.B. and released in 1998. It involves controlling a ball within a world that contains a series of objects. The world consists of a \emph{platform} on which the ball can move.

An object can be a coin, a piece of fruit or a key, each of which are \emph{collected} when a player moves into them. The ball is controlled through the use of the \emph{directional pad} and the \emph{cross} button shown in Figure \ref{fig:controller}.
Pressing the right or left directional button rotates the direction of view 90 degrees clockwise or anti-clockwise about the line perpendicular to the platform.
Pressing the up directional button moves the player forwards on the platform, in the direction that the camera is facing. Pressing the cross button makes the ball jump, this can be pressed simultaneously with the forward directional button to make the ball \emph{jump forwards}.
Jumping forwards moves the player two squares forwards, over the square in front of it.
If the player jumps onto a square that doesn't exist, the game will end.

\subsection{Actions}
The definition of the action space for Kula World is relatively simple. We omitted the \emph{jump} action since this served no purpose within the levels that were tested. In total, the action space is given by:
\begin{equation}
    A = \left\{ \textsc{Forward}, \quad\textsc{LookRight},\quad \textsc{LookLeft},\quad \textsc{JumpForward} \right\}
\end{equation}

There is a clock on each level, which counts down in seconds from a level-specific start value. To complete a level, the player must pick up all of the keys and move to a \emph{goal square} before the clock reaches zero. Collecting objects gains points, which are added to the player's \emph{score} for the level.

It is not suitable to employ frame skipping in Kula World, since moves can vary in length. A jump, for example, takes roughly a second longer than a camera rotation. The duration of moves can also depend on the specific state of a level; for example, moving forwards to collect a coin takes longer than moving forwards without collecting a coin. This is a problem since a move cannot be carried out while another is taking place. Further, since the logic for the game takes place within the CPU of the console, it is not possible in general to predict the duration of a move prior to it finishing. Instead, the asynchronous approach described earlier is used. We did not use any kind of frame stacking since a snapshot of the console's display does not contain any ambiguous motion.

\subsection{Rewards}

There are several ways of `losing the game' in Kula World: falling off the edge of the platform, being `spiked' by an object in the game or running out of time. We consider these to be identical events in terms of their reward, although the abstraction supports assigning different values to each way of losing. It also supports adding the player's \emph{remaining time} to the state encoding, to ensure that agents aren't misled by time running out in the level. If the remaining time of a level is low, agents will learn that moves are likely to yield negative rewards and modify their behaviour appropriately. A constant negative value is added to the reward incurred by all actions in order to discourage agents from making moves that do not lead to timely outcomes.

A user of this abstraction specifies a function \texttt{score\_to\_reward}, which takes a \emph{score change} (such as 250 for a coin, 1000 for a key and 0 for a non-scoring action) and returns the instantaneous reward. In addition, they specify fixed rewards for winning and losing the game. While they can choose \texttt{score\_to\_reward} arbitrarily, most implementations will ensure that: the instantaneous reward increases for increasing changes in score, the reward of a non-scoring action is a small negative value and the \emph{discounted} sum of these rewards is always bounded, to aid stability in learning.

\subsection{State}

\begin{figure}[h]
\vspace{-0.2cm}
\centerline{\includegraphics[width=\textwidth,keepaspectratio]{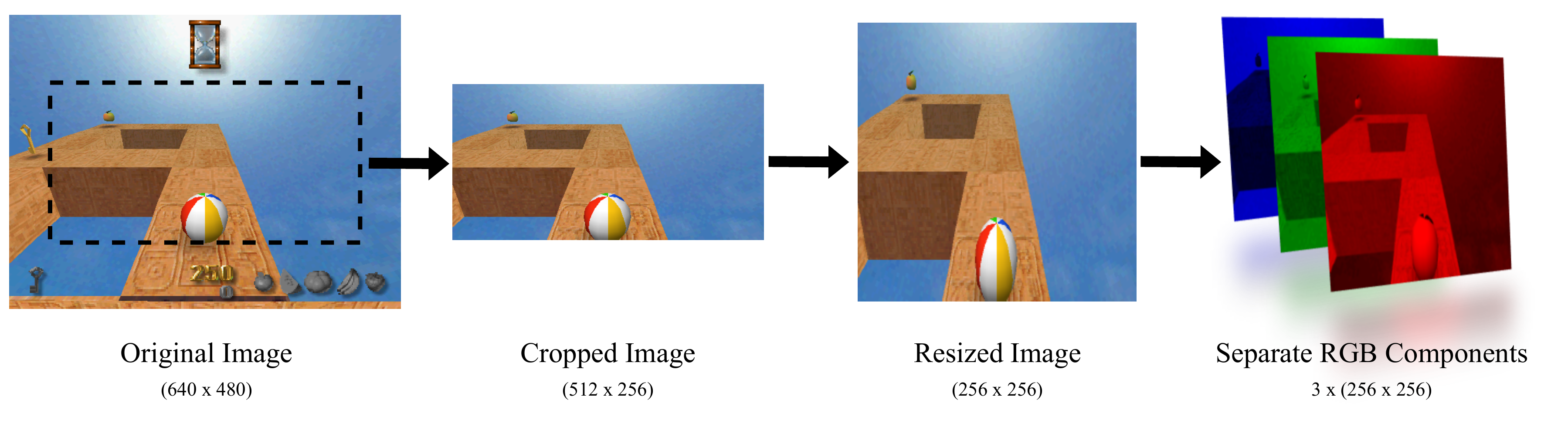}}
\vspace{-0.2cm}
\caption{Frame processing used for complex state representations. Images have their red, green and blue (RGB) channels separated.}
\label{fig:frameprocess}
\vspace{-0.4cm}
\end{figure}

The abstraction does not prescribe a state encoding, instead it returns a tuple of relevant data after each move has finished. The contents of the tuple are:
\begin{itemize}
    \item \texttt{visual}: an RGB array derived using the process shown in Figure \ref{fig:frameprocess}.
    \item \texttt{reward}: the value of instantaneous reward resulting from the move that has just been executed.
    \item \texttt{playing}: a value indicating whether the player is still `alive'.
    \item \texttt{clock}: the number of seconds that remain in which the player must complete the level.
    \item \texttt{sound}: one of either: \texttt{None}, if the practitioner has not instructed the abstraction to record the sound of moves; an array of \emph{Mel-frequency Cepstral Coefficients} (MFCCs), if the abstraction was instantiated with \texttt{use\_mfcc} set to \texttt{True}; or an array describing the raw audio output of the console over the duration of the move, otherwise.
    \item \texttt{duration\_real}: the amount of time the move took to complete.
    \item \texttt{duration\_game}: the amount of the player's remaining time that the move took to complete, relative to the in-game clock.
    \item \texttt{score}: the score that the player has achieved so far \emph{in the current episode}.
\end{itemize}
When a move does not make a sound, \texttt{sound} will be an empty array. When a move makes a sound with a duration \emph{shorter} than that of the move, silence from the recording will be removed from both its start and end. If a move's sound lasts \emph{longer} than the duration of the move, the game will continue running until either the audio has ceased \emph{or} the maximum recording time has been exceeded. The result of these features is that the audio output of moves is succinct, as shown in Appendix \ref{apdx:audio}, but that moves may take longer to execute when the abstraction is recording audio.

\section{Evaluation}

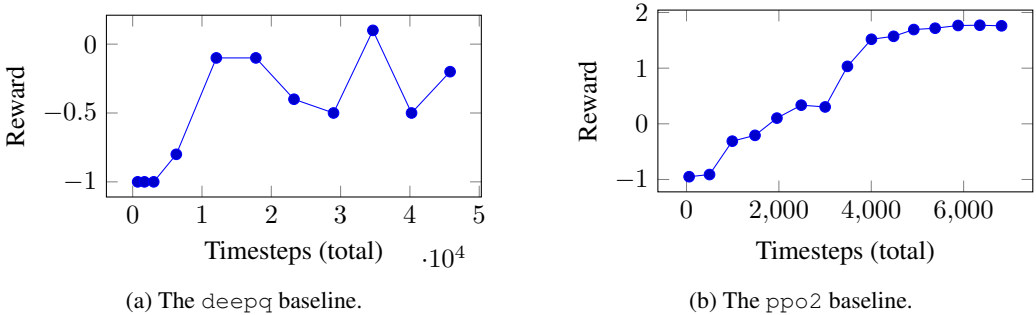
\begin{figure}[h]
\vspace{-0.4cm}
  \centering
\begin{subfigure}[b]{.47\textwidth}
  \centering
    \begin{tikzpicture}
            \begin{axis}[width=\textwidth,
                xlabel={Timesteps (total)},
                ylabel={Reward},
                height=4cm,
                legend style={at={(0.5,-0.3)},anchor=north}]
                \addplot table [x=steps, y=meanepreward, col sep=comma] {Baselines/deepq_output.csv};
            \end{axis}
    \end{tikzpicture}
    \caption{The \texttt{deepq} baseline.}
    \label{fig:deepqplot}
\end{subfigure}\hfill
\begin{subfigure}[b]{.47\textwidth}
  \centering
    \begin{tikzpicture}
            \begin{axis}[width=\textwidth,
                xlabel={Timesteps (total)},
                ylabel={Reward},
                height=4cm,
                legend style={at={(0.5,-0.3)},anchor=north}]
                \addplot table [x=misc/time_elapsed, y=eprewmean, col sep=comma] {Baselines/ppo_output.csv};
            \end{axis}
    \end{tikzpicture}
    \caption{The \texttt{ppo2} baseline.}
    \label{fig:ppoplot}
\end{subfigure}
    \caption{A plot of the average reward over 100 episodes, as given by OpenAI Baselines, against the number of timesteps that the agent had played. The agents were trained on Level 1 of Kula World using a visual state encoding.}
    \label{fig:evalsimples}
\end{figure}

Appendix \ref{apdx:usage} gives example usage of both the game abstraction and the OpenAI Gym interface. The \texttt{Kula-v1} Gym environment supports Kula World levels from 1 up to 10, uses the console's screen as its state encoding and (by default) uses the reward function described in Table \ref{table:reward}. This environment was used with \texttt{deepq} and \texttt{ppo2} from OpenAI Baselines\footnote{Available at \url{https://github.com/openai/baselines}}. The \texttt{ppo2} baseline is an implementation of \emph{Proximal Policy Optimization Algorithms} from Schulman \emph{et al.} \cite{PPO}. The results of these are shown in Figures \ref{fig:deepqplot} and \ref{fig:ppoplot}.

\begin{table}[h]
  \vspace{-0.5cm}
  \caption{Reward function.}
  \label{table:reward}
  \centering
  \begin{tabular}{lll}
    \toprule
    Event     & Score change     & Reward \\
    \midrule
    Coin collect & +250 & 0.2 \\
    Key collect & +1000 & 0.4 \\
    Fruit collect & +2500 & 0.6 \\
    Win level    & - & 1 \\
    Lose level    & - & -1 \\
    \bottomrule
  \end{tabular}
\end{table}

\begin{figure}[ht]
    \centering
\begin{subfigure}{.55\textwidth}
        \centering
        \begin{subfigure}[t]{.20\textwidth}
          \centering
          \includegraphics[width=\textwidth,keepaspectratio]{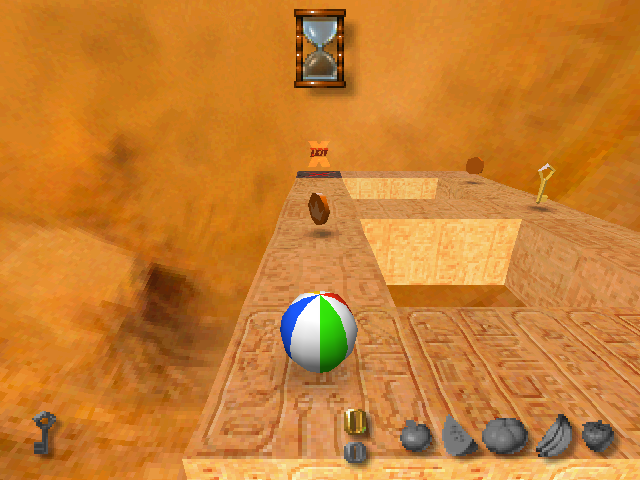}
        \end{subfigure}
        \begin{subfigure}[t]{.20\textwidth}
          \centering
          \includegraphics[width=\textwidth,keepaspectratio]{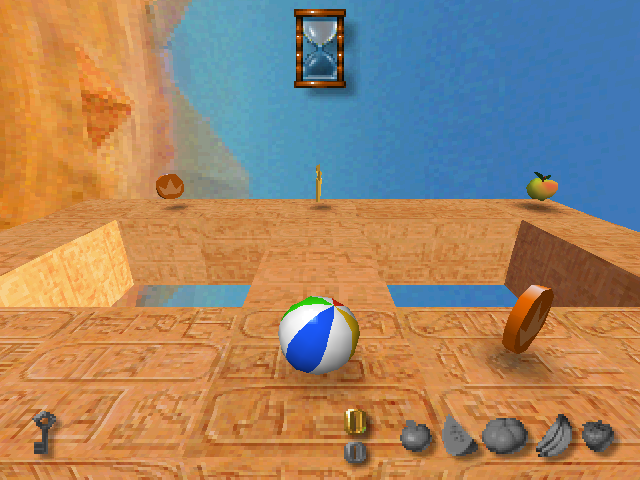}
        \end{subfigure}
        \begin{subfigure}[t]{.20\textwidth}
          \centering
          \includegraphics[width=\textwidth,keepaspectratio]{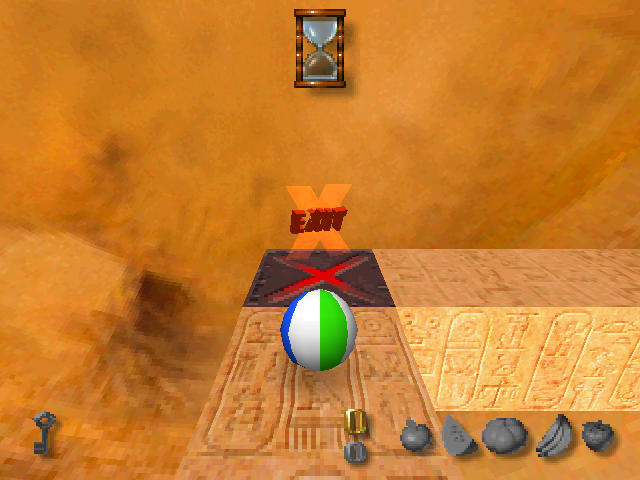}
        \end{subfigure}
        \begin{subfigure}[t]{.20\textwidth}
          \centering
          \includegraphics[width=\textwidth,keepaspectratio]{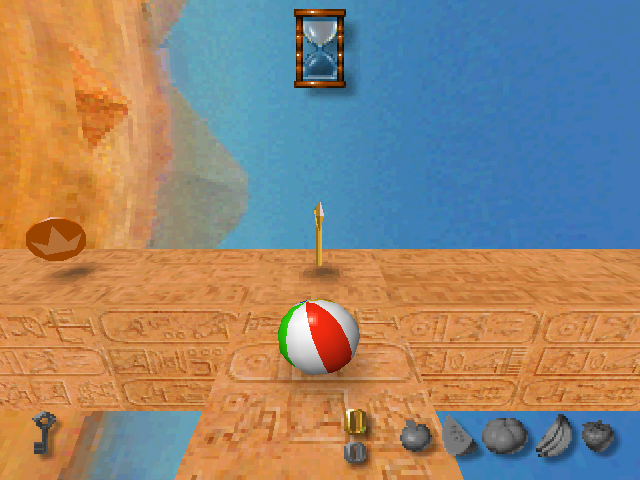}
        \end{subfigure}

        \begin{subfigure}[t]{.20\textwidth}
          \centering
          \includegraphics[width=\textwidth,keepaspectratio]{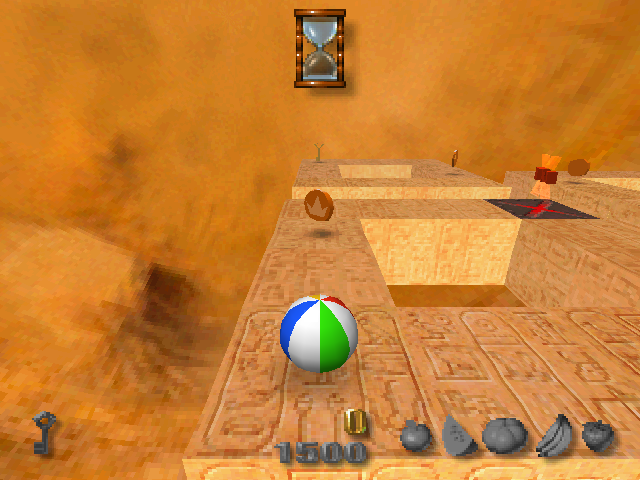}
        \end{subfigure}
        \begin{subfigure}[t]{.20\textwidth}
          \centering
          \includegraphics[width=\textwidth,keepaspectratio]{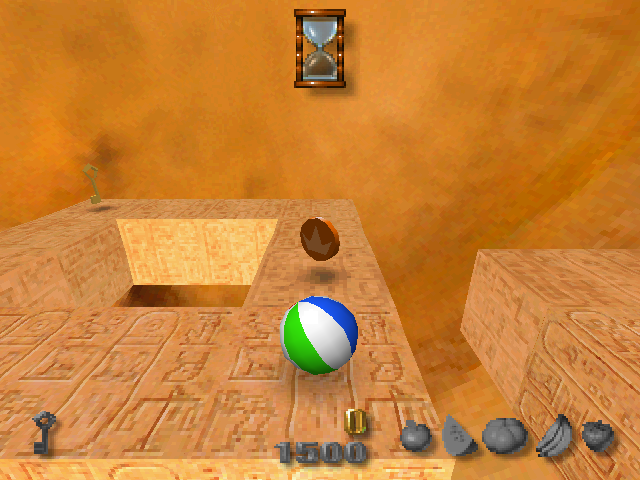}
        \end{subfigure}
        \begin{subfigure}[t]{.20\textwidth}
          \centering
          \includegraphics[width=\textwidth,keepaspectratio]{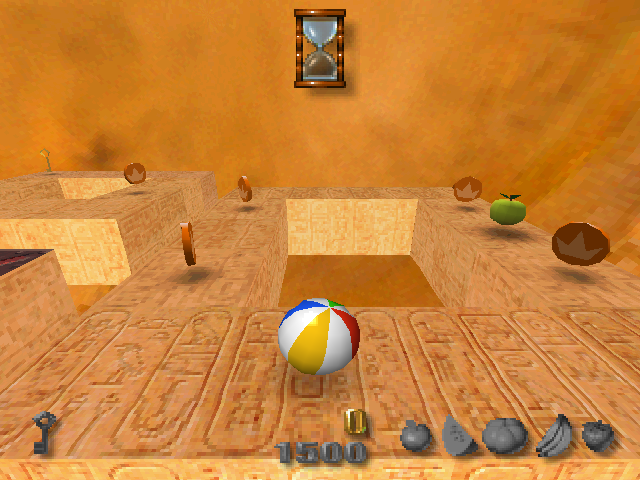}
        \end{subfigure}
        \begin{subfigure}[t]{.20\textwidth}
          \centering
          \includegraphics[width=\textwidth,keepaspectratio]{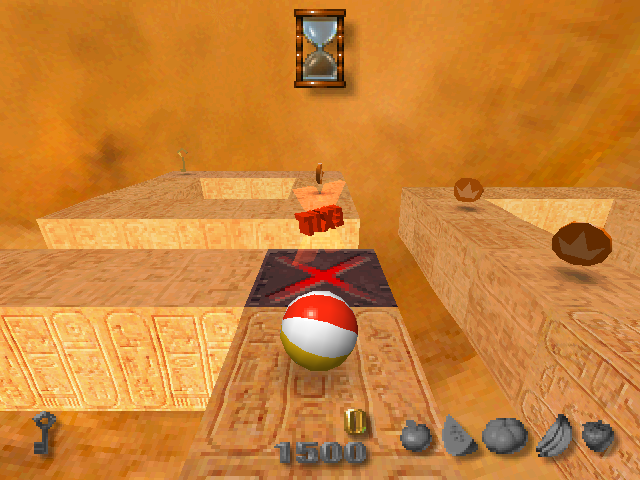}
        \end{subfigure}

        \begin{subfigure}[t]{.20\textwidth}
          \centering
          \includegraphics[width=\textwidth,keepaspectratio]{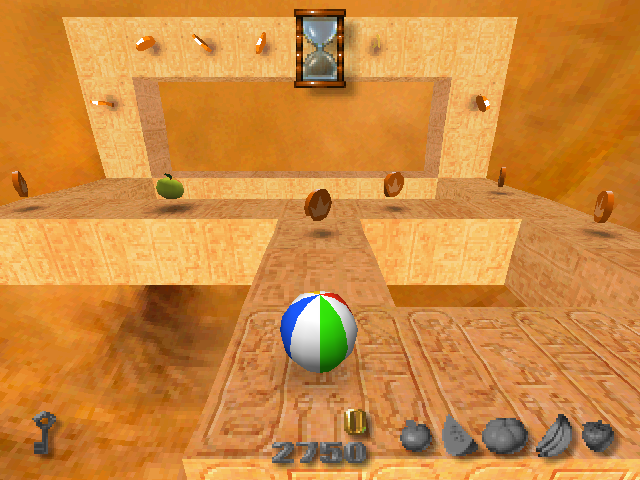}
        \end{subfigure}
        \begin{subfigure}[t]{.20\textwidth}
          \centering
          \includegraphics[width=\textwidth,keepaspectratio]{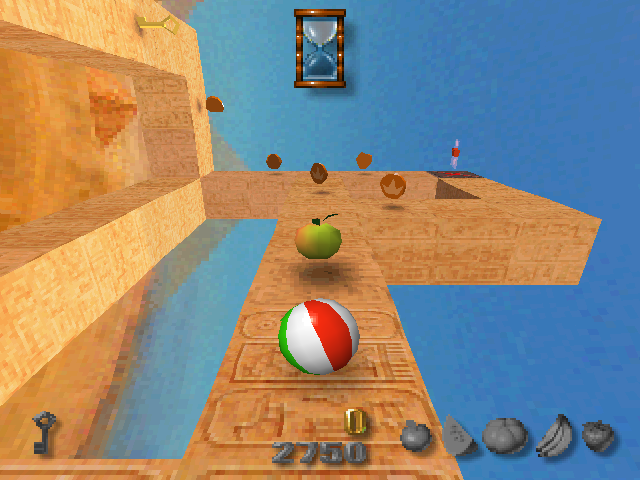}
        \end{subfigure}
        \begin{subfigure}[t]{.20\textwidth}
          \centering
          \includegraphics[width=\textwidth,keepaspectratio]{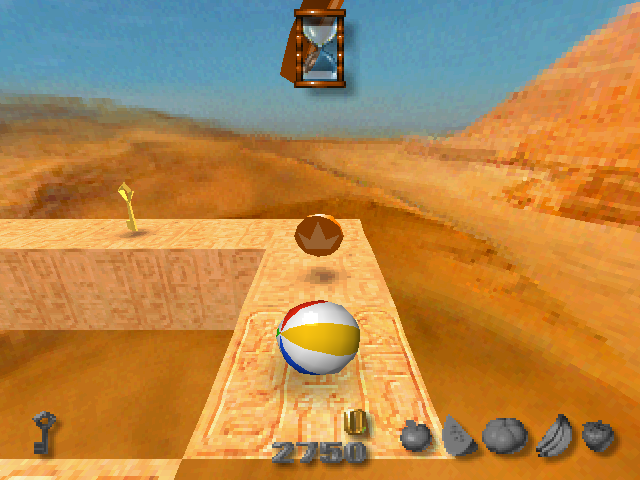}
        \end{subfigure}
        \begin{subfigure}[t]{.20\textwidth}
          \centering
          \includegraphics[width=\textwidth,keepaspectratio]{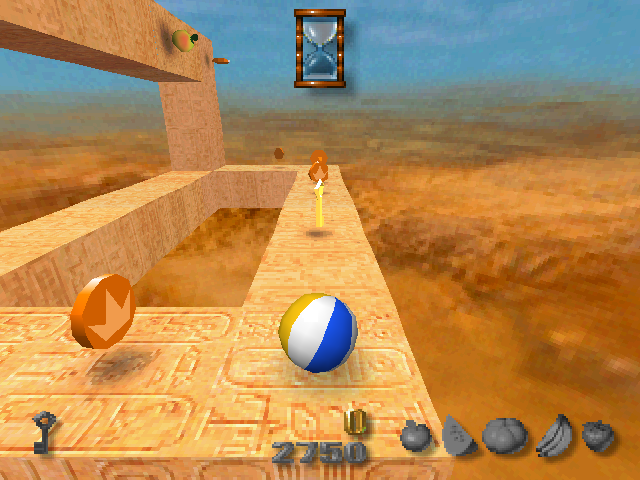}
        \end{subfigure}
        
        \caption{Starting locations, as used in training.}
\end{subfigure}\hfill
\begin{subfigure}{.4\textwidth}
    \centering
    \includegraphics[width=.9\textwidth,keepaspectratio]{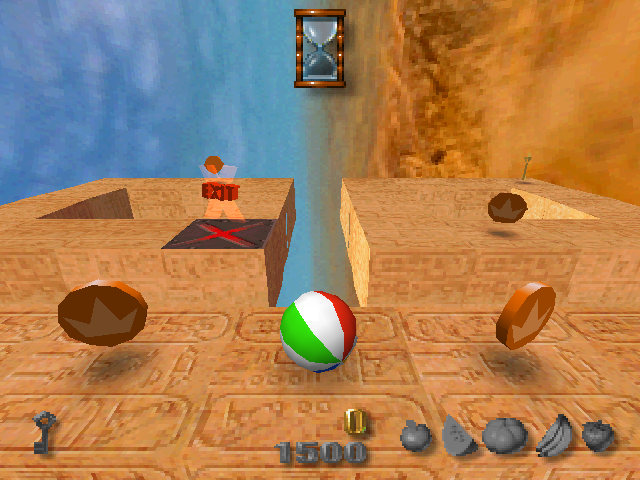}
    \caption{The evaluation episode start location.}
\end{subfigure}
\caption{Screenshots from Kula World showing each possible start position.}
\label{fig:allstates}
\vspace{-0.2cm}
\end{figure}

This approach to training is not particularly interesting, since the agent is presented with the same situation in each episode. As a result, the agent simply learns a set of moves which will result in it winning \emph{that} level in the shortest time and with the highest score. To make the problem more challenging, we propose varying the player's start position within the level and the use of multiple levels. The game does not offer this feature by default, but it can be implemented using the tools available in PSXLE. To add an additional starting position for a player, simply save a PSXLE \texttt{state} with it at the desired position. We chose to limit the time available to complete each level to 80 seconds, rather than the usual 100, so that each starting state allowed the same amount of time to complete the level. The environment \texttt{Kula-random-v1} takes this approach. It presents the agent with one of four different starting position in each of the first three levels of the game. There is an additional starting position within Level 2 that is reserved for use in \emph{validation episodes} in which no learning takes place and the agent picks moves from its learned policy. Each of the starting positions are shown in Figure \ref{fig:allstates}.

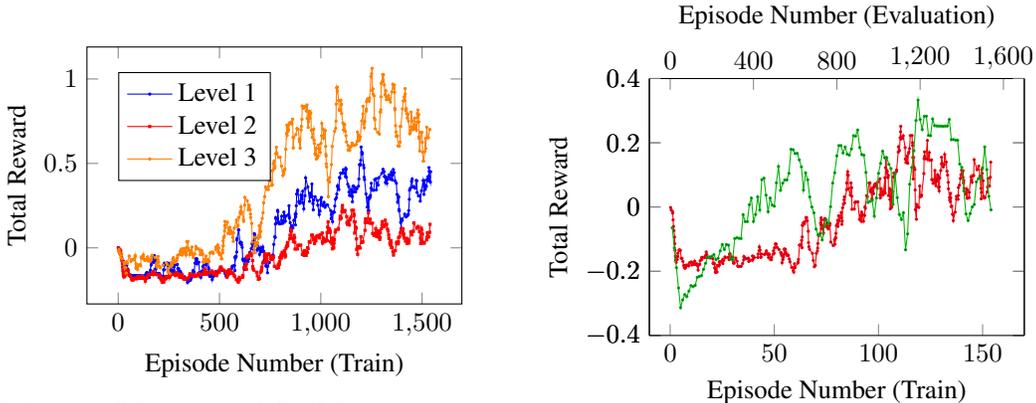
\begin{figure}[h]
\vspace{-0.3cm}
  \centering
\begin{subfigure}[b]{.47\textwidth}
  \centering
    \begin{tikzpicture}
        \begin{axis}[width=\textwidth,
            height=5cm,
            legend style={at={(0cm,2.8cm)},anchor=north west},
            xlabel={Episode Number (Train)},
            ylabel={Total Reward}]
            
            \addplot+[blue,
            mark size=0.4pt,
            mark options={fill=blue}] table [x=X, y=Y, col sep=comma] {data/lv1_moving_average_mix.csv};
            
            \addplot+[red,
            mark size=0.4pt,
            mark options={fill=red}] table [x=X, y=Y, col sep=comma] {data/lv2_moving_average_mix.csv};
            
            \addplot+[orange,
            mark size=0.4pt,
            mark options={fill=orange}] table [x=X, y=Y, col sep=comma] {data/lv3_moving_average_mix.csv};
            
            \addlegendentry{Level 1}
            \addlegendentry{Level 2}
            \addlegendentry{Level 3}
        \end{axis}
        
    \end{tikzpicture}
    \caption{Agent proficiency on each level during training. Start locations within each level are chosen uniformly at random from those shown in Figure \ref{fig:allstates}. The line shows a moving average over 10 training episodes.}
    \label{fig:eval:genr_compare}
\end{subfigure}\hfill
\begin{subfigure}[b]{.47\textwidth}
  \centering
    \begin{tikzpicture}
        \begin{axis}[width=\textwidth,
            height=5cm,
            legend style={at={(0.3cm,2cm)},anchor=north west},
            xlabel={Episode Number (Evaluation)},
            axis x line*=top,
            xmin=-100,
            xmax=1700,
            ymin=-0.4,
            ymax=0.4, xtick distance=400]

            \addplot+[red,
            mark size=0.4pt,
      yticklabels={,,}] table [x=X, y=Y, col sep=comma] {data/lv2_moving_average_mix.csv};
        \end{axis}

        \begin{axis}[width=\textwidth,
            height=5cm,
            legend style={at={(0.3cm,2cm)},anchor=north west},
            xlabel={Episode Number (Train)},
            ylabel={Total Reward},
            axis x line*=bottom,
            xmin=-10,
            xmax=170,
            ymin=-0.4,
            ymax=0.4]
            
            \addplot+[black!30!green,
            mark size=0.4pt] table [x=X, y=Y, col sep=comma] {data/test_moving_average_mix.csv};
        \end{axis}
    \end{tikzpicture}
    \caption{The green line shows a moving average over 5 validation episodes for the reserved location, the red line is the same as in Figure \ref{fig:eval:genr_compare}.}
    \label{fig:eval:genr_test}
\end{subfigure}
    \caption{Evaluation of a simple DQN implementation for \texttt{Kula-random-v1}.}
    \label{fig:evalrands}
\end{figure}

We implemented a simple DQN agent based on the network presented in Mnih \emph{et al.}\cite{NatureDeepMind}, for use with \texttt{Kula-random-v1}. The results are summarised in Figure \ref{fig:eval:genr_compare} and Figure \ref{fig:eval:genr_test}.

Surprisingly, the agent was most successful in learning how to play Level 3, despite it being the most complex for human players. This may be due to Level 3 introducing new physics that are not intuitive to humans, but are easier to navigate for the agent. Level 2 requires players jump between islands, meaning a jump can either be very profitable for a player (in that it gains access to more coins) or very bad (because the player falls off the platform). This discrepancy means that agents must disambiguate a jump that is required for it to reach the goal square from one that will end the game. By comparison, Level 3 does not require any jumps to complete.

The agent performs similarly from the reserved start position as it did from the other available locations in Level 2. It is promising that the agent did not perform notably \emph{worse} when given an unseen start state, it is also worth noting that the performance of the agent was fairly lacklustre on Level 2 in general. 

In order to leverage the audio recording features of PSXLE, we developed another environment: \texttt{Kula-audio-v1}. This environment includes a representation of the audio output from each move within its state encoding. The state encoding is the richest yet, containing visual output, audio output, time remaining on the level and score. Since MFCCs have been successful in machine learning contexts for classifying audio data, they are used here to represent the audio output of a move. The MFCC outputs for some moves are shown in Appendix \ref{apdx:audio}.

Finally, since the agent starts each episode from a specified state, the states in its memory are dominated by those that occur at the start of the level. For PlayStation games, which are usually quite complex and have long levels, this can result in lenghtly training times. Prioritised experience replay \cite{PExperienceReplay} attempts to combat this by replaying state transitions that the agent learned the most from. We propose a modification to this, \emph{prioritised state replay}, in which an agent can \emph{resume play} from a state which it has previously visited, allowing the agent to perform a different action. This can be trivially implemented using the \texttt{save\_state} and \texttt{load\_state} methods in PSXLE and could help to reduce the time required to learn complex games. No agents have been trained using this approach.

\vspace{-0.5cm}
\section{Conclusion}

The results from \texttt{deepq} and \texttt{ppo2} baselines show how two different RL algorithms can perform to a vastly different standard within the Kula World environment. From this, it is clear that PlayStation games can represent suitable environments in which to evaluate the effectiveness of RL algorithms.

The approach shown in this paper for Kula World can be customised to work with many different PlayStation games. The design of PSXLE makes it simple to build abstractions that support interfaces like OpenAI Gym. Armed with such abstractions, researchers will be able to apply existing RL implementations on more complex environments with richer state spaces. The challenges that the audio and complex visual rendering in PlayStation games present to RL could help us to close the gap between what we want agents to achieve and the methodology that we use to evaluate them.

\pagebreak

\begin{appendices}

\section{Library usage}
\label{apdx:usage}

To demonstrate the functionality of the work presented in this paper, example scripts are shown below. These are chosen to show how practitioners can interact with the environment at each level of abstraction.

\subsection{Console}
This script creates a console running Kula World, loads Level 1 and moves the player forwards two steps.

\begin{lstlisting}[language=Python]
import psxle

# Make a console running kula world...
c = psxle.Console("~/.psxle/isos/kula.iso", display=psxle.Display.NONE)
c.load("level1")
c.run()

# Press up twice, with a gap of 100ms
c.touch_button(psxle.Control.UP)
c.delay_button(100)
c.touch_button(psxle.Control.UP)
\end{lstlisting}

\subsection{Game abstraction}
This script creates a \texttt{Game} object for Kula World and plays it randomly.

\begin{lstlisting}[language=Python]
from psxle import Display
import random
from gym_kula.kula import Game
from gym_kula.kula import State

# make a game at Level 1, which runs at high fps
game = Game("level1", display = Display.NONE, fast = True)
while True:
    g.play()
    # while the player is still alive
    while g.state == State.PLAYING:
        possible_moves = g.getMoveOptions()
        # pick a move
        move = random.choice(possible_moves)
        outcome = g.move(move)
    # Game Over...
    print("Outcome:", g.interpretState())
\end{lstlisting}

\subsection{Interface}
This script uses the OpenAI Gym implementation of Kula World to play randomly. Note that the methods being used are entirely game-agnostic.

\begin{lstlisting}[language=Python]
import gym
import gym_kula

kula = gym.make("Kula-v1")
while True:
    kula.reset()
    done = False
    while not done:
        # pick an action at random
        action = kula.action_space.sample()
        screen, reward, done, info = kula.step(action)
        # screen has display output
        # reward has the reward gained in the move
        # done is True if the game has ended
        # info includes duration of move, score, time left
\end{lstlisting}

\section{Audio outputs}
\label{apdx:audio}

\begin{figure}[h]
    \centering
    \begin{subfigure}[t]{.48\textwidth}
      \centering
      \includegraphics[width=\textwidth,keepaspectratio]{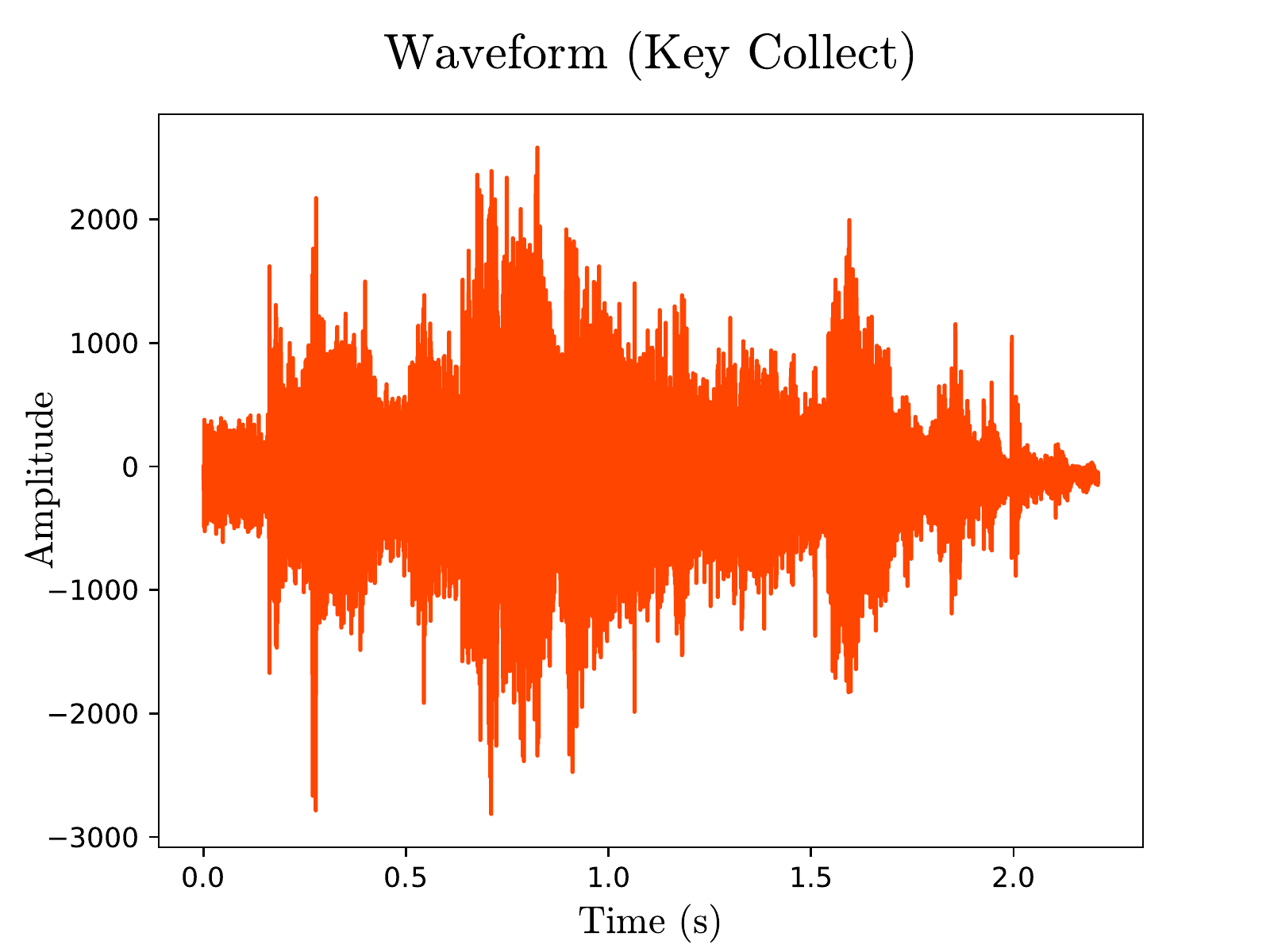}
    \end{subfigure}\hfill
    \begin{subfigure}[t]{.48\textwidth}
      \centering
      \includegraphics[width=\textwidth,keepaspectratio]{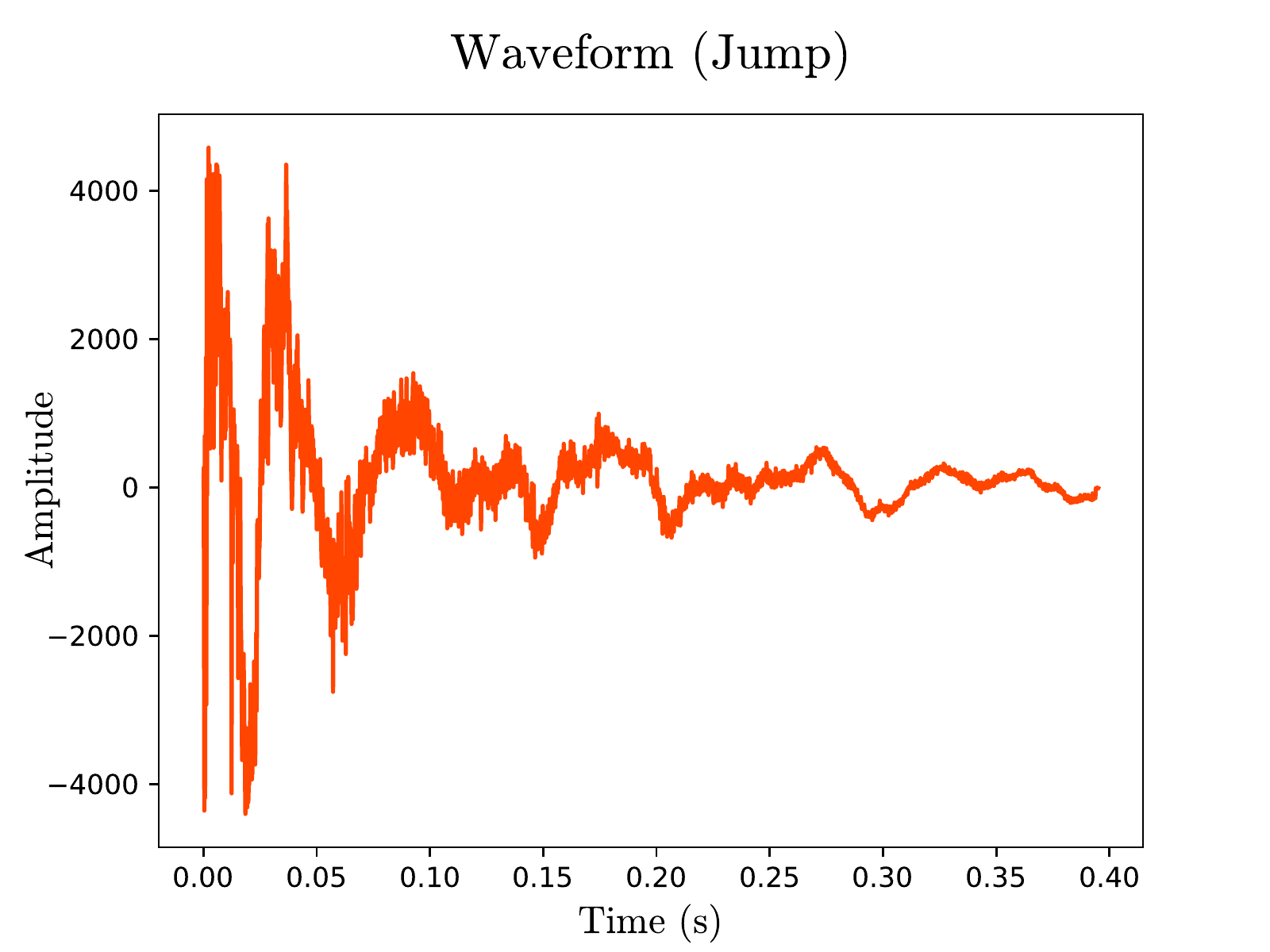}
    \end{subfigure}
      \\
    \begin{subfigure}[t]{.48\textwidth}
      \centering
      \includegraphics[width=\textwidth,keepaspectratio]{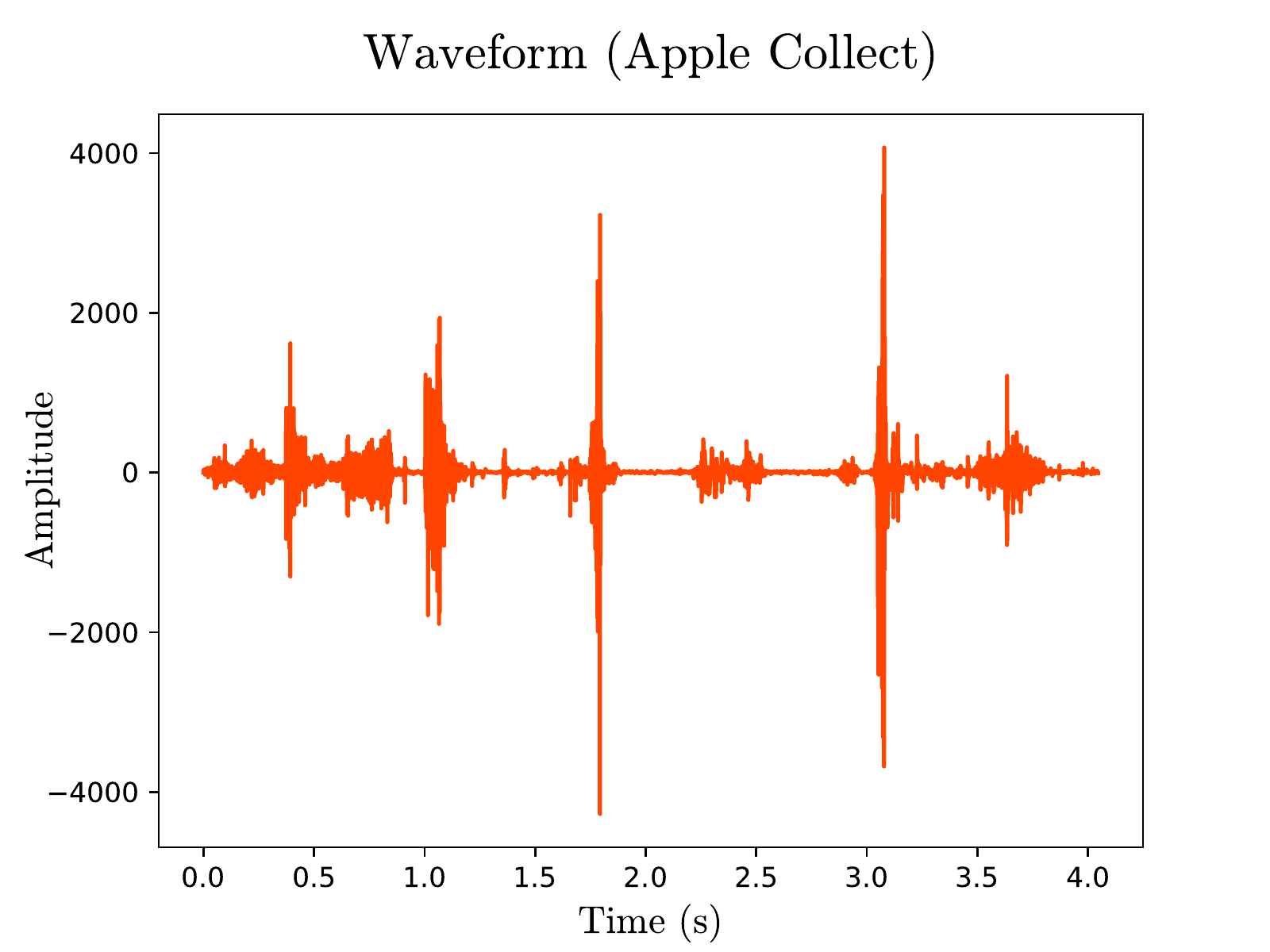}
    \end{subfigure}\hfill
    \begin{subfigure}[t]{.48\textwidth}
      \centering
      \includegraphics[width=\textwidth,keepaspectratio]{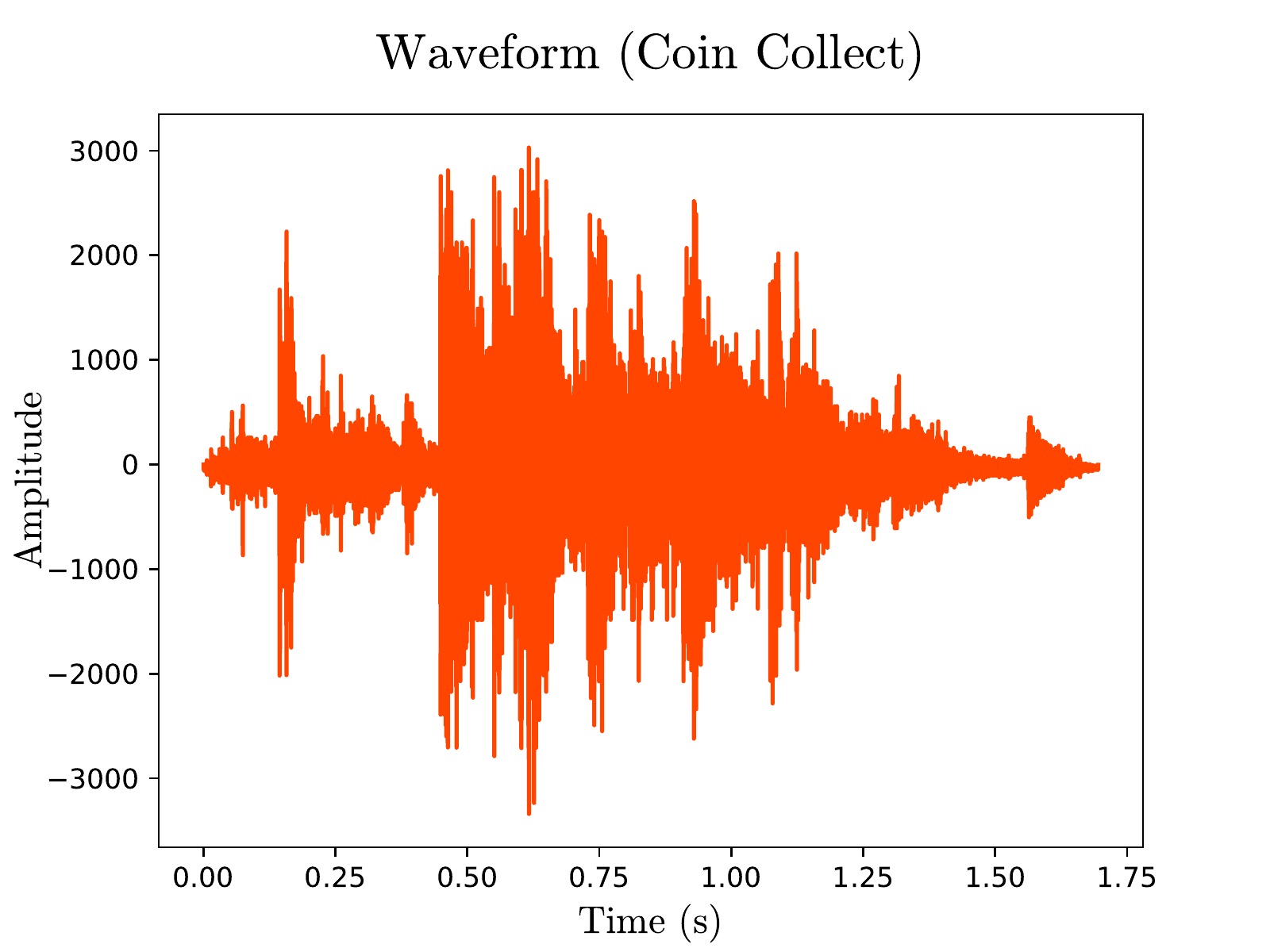} 
      \end{subfigure}
    
    \caption{Raw audio output for a selection of moves within Kula World, obtained using PSXLE's audio recording feature. The abstraction employs cropping and extended move times to ensure that the full audio waveform for each move is captured.}
    \label{fig:imp:waveforms}
\end{figure}

\begin{figure}
    \centering
    \begin{subfigure}[t]{.48\textwidth}
      \centering
      \includegraphics[width=\textwidth,keepaspectratio]{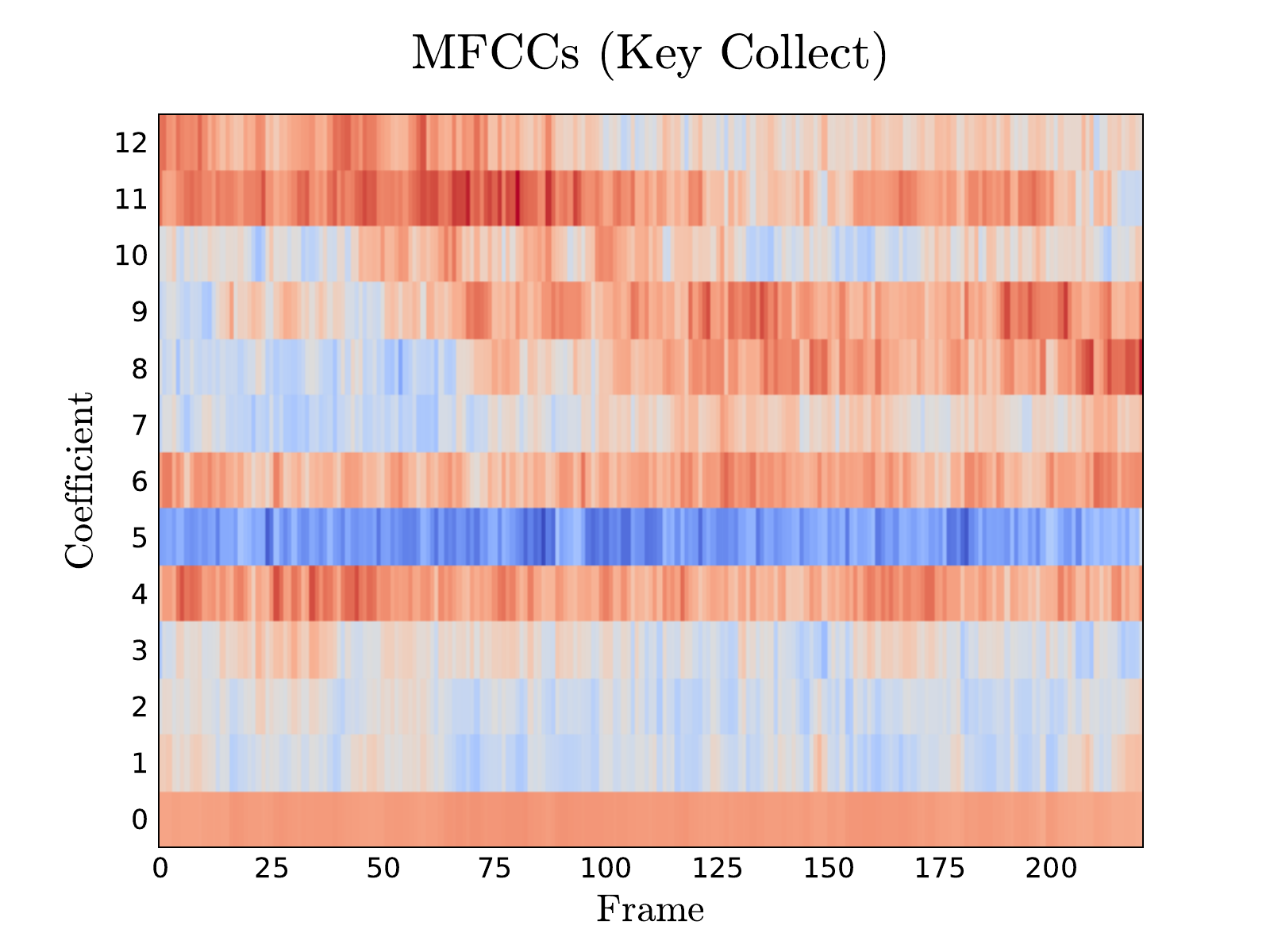}
    \end{subfigure}\hfill
    \begin{subfigure}[t]{.48\textwidth}
      \centering
      \includegraphics[width=\textwidth,keepaspectratio]{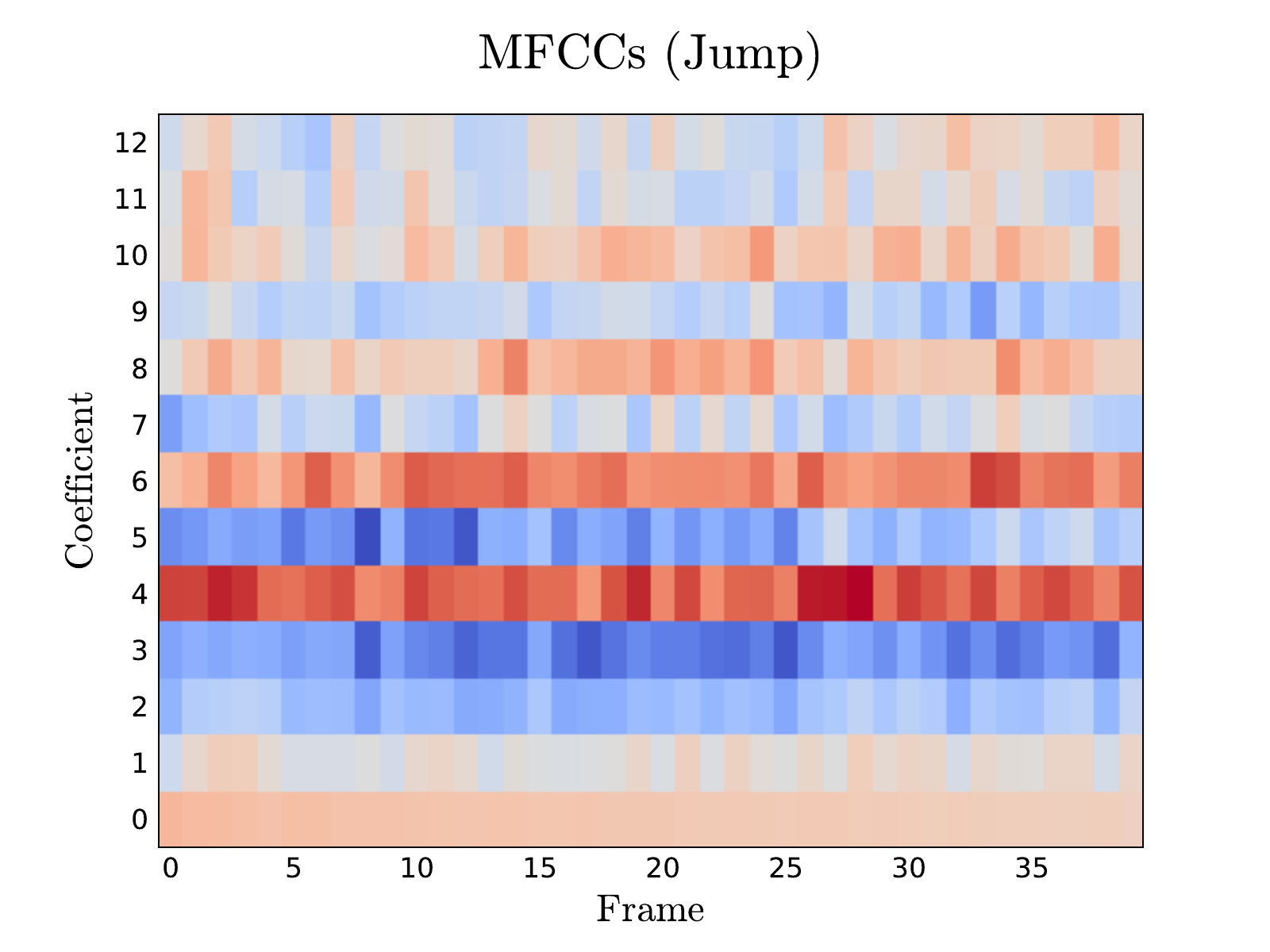} \\
    \end{subfigure} \\
    \begin{subfigure}[t]{.48\textwidth}
      \centering
      
      \includegraphics[width=\textwidth,keepaspectratio]{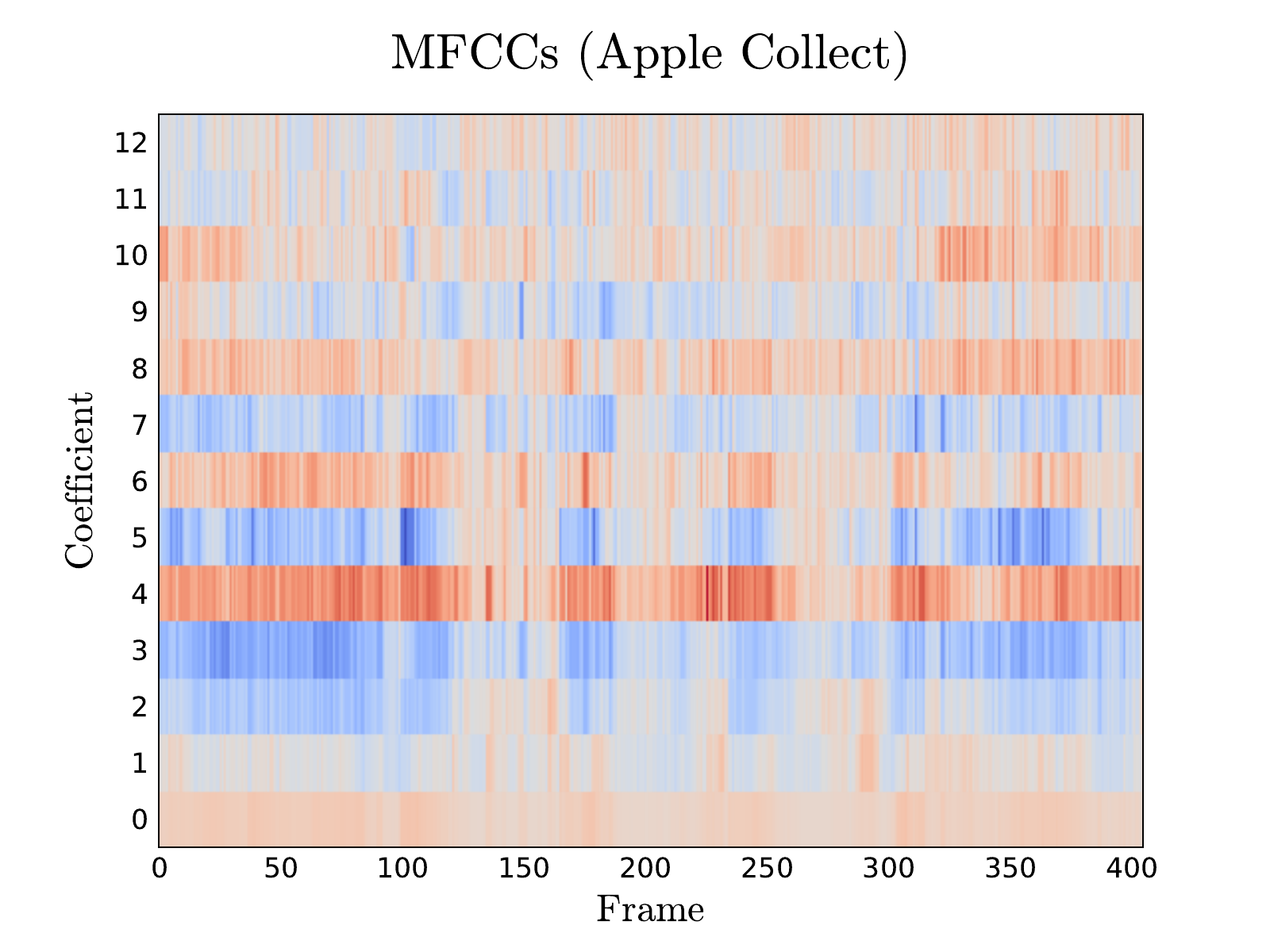}
    \end{subfigure}\hfill
    \begin{subfigure}[t]{.48\textwidth}
      \centering
      \includegraphics[width=\textwidth,keepaspectratio]{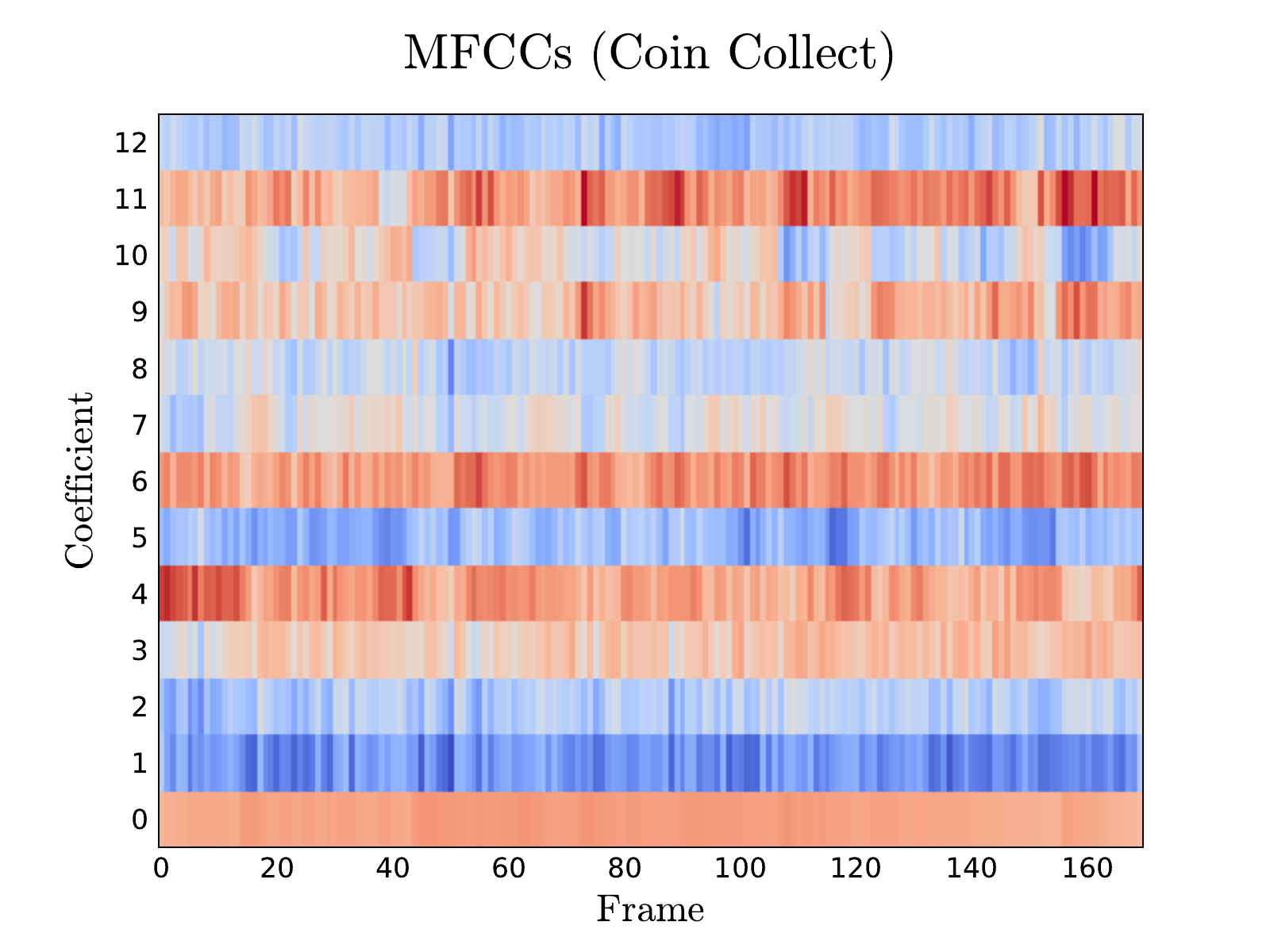} 
      \end{subfigure}
    \caption{MFCCs derived from the waveforms in Figure \ref{fig:imp:waveforms}.}
    \label{fig:imp:mfccs}
\end{figure}

\FloatBarrier
\end{appendices}

\pagebreak

\bibliography{refs}

\end{document}